\documentclass[runningheads]{llncs}

\usepackage{eccv}

\usepackage{eccvabbrv}

\usepackage{epsfig}
\usepackage{graphicx}

\usepackage{comment}

\usepackage{pifont}

\usepackage{microtype}
\usepackage{arydshln}
\usepackage{tabularx}
\usepackage{adjustbox}
\usepackage{array}
\usepackage{booktabs}
\usepackage{colortbl}
\usepackage{float,wrapfig}
\usepackage{hhline}
\usepackage{multirow}
\usepackage[percent]{overpic}

\usepackage{stmaryrd}
\usepackage{breqn}

\usepackage{amsfonts}

\usepackage{bm}
\usepackage{nicefrac}
\usepackage{microtype}
\usepackage{dsfont}
\usepackage{changepage}
\usepackage{extramarks}
\usepackage{fancyhdr}
\usepackage{setspace}
\usepackage{soul}
\usepackage{hhline}
\usepackage{algorithmicx}
\usepackage{algorithm}
\usepackage{algpseudocode}
\usepackage{amssymb}
\usepackage{multirow}

\usepackage{makecell}

\usepackage{enumitem}
\usepackage[title]{appendix}

\usepackage{placeins}

\makeatletter
\newcommand{\smallsym}[2]{#1{\mathpalette\make@small@sym{#2}}}
\newcommand{\make@small@sym}[2]{%
  \vcenter{\hbox{$\m@th\downgrade@style#1#2$}}%
}
\newcommand{\downgrade@style}[1]{%
  \ifx#1\displaystyle\scriptstyle\else
    \ifx#1\textstyle\scriptstyle\else
      \scriptscriptstyle
  \fi\fi
}
\makeatother

\newcommand{\ignorethis}[1]{}

\newcommand{\myparagraph}[1]{\vspace{1pt} \noindent \textbf{#1} \ }

\def\1{\bm{1}}

\newcolumntype{L}[1]{>{\raggedright\let\newline\\\arraybackslash\hspace{0pt}}m{#1}}
\newcolumntype{C}[1]{>{\centering\let\newline\\\arraybackslash\hspace{0pt}}m{#1}}
\newcolumntype{R}[1]{>{\raggedleft\let\newline\\\arraybackslash\hspace{0pt}}m{#1}}

\newcommand{\ignore}[1]{}

\renewcommand*{\thefootnote}{\arabic{footnote}}

\makeatletter
\DeclareRobustCommand\onedot{\futurelet\@let@token\@onedot}
\def\@onedot{\ifx\@let@token.\else.\null\fi\xspace}
\def\eg{e.g\onedot,\xspace} 

\def\ie{i.e\onedot,\xspace}

\def\etc{\emph{etc}\onedot} 
 
\def\etal{\emph{et al}\onedot}
\makeatother

\usepackage{color}

\usepackage{bbm}
\usepackage{bbding}
\definecolor{mygray}{gray}{.9}

\usepackage{arydshln}
\makeatletter
\def\adl@drawiv#1#2#3{%
        \hskip.5\tabcolsep
        \xleaders#3{#2.5\@tempdimb #1{1}#2.5\@tempdimb}%
                #2\z@ plus1fil minus1fil\relax
        \hskip.5\tabcolsep}
\newcommand{\cdashlinelr}[1]{%
  \noalign{\vskip\aboverulesep
           \global\let\@dashdrawstore\adl@draw
           \global\let\adl@draw\adl@drawiv}
  \cdashline{#1}
  \noalign{\global\let\adl@draw\@dashdrawstore
           \vskip\belowrulesep}}
\makeatother

\usepackage[accsupp]{axessibility}  

\usepackage{hyperref}

\usepackage{orcidlink}

\begin{document}

\title{Learn from the Learnt: Source-Free Active
Domain Adaptation via Contrastive Sampling and Visual Persistence} 


\author{Mengyao Lyu\inst{1,2}$^\star$ \and
Tianxiang Hao\inst{1,2}$^\star$ \and
Xinhao Xu\inst{1,2} \and
Hui Chen\inst{1,2}\textsuperscript{\Envelope} \and
Zijia Lin\inst{1} \and \\
Jungong Han\inst{1,2} \and
Guiguang Ding\inst{1,2}\textsuperscript{\Envelope}
}

\authorrunning{M.~Lyu et al.}

\institute{Tsinghua University, Beijing, China \and BNRist, Beijing, China \\
\email{
mengyao.lyu@outlook.com, 
beyondhtx@gmail.com,
\{xxh22@mails, huichen@mail\}.tsinghua.edu.cn,
linzijia07@tsinghua.org.cn, 
jungonghan77@gmail.com, 
dinggg@tsinghua.edu.cn
}}

\maketitle

\renewcommand{\thefootnote}{$\star$}
\footnotetext[2]{Equal Contribution \;\; \Envelope\ Corresponding Author}

\renewcommand{\thefootnote}{\arabic{footnote}}

\begin{abstract}
    Domain Adaptation (DA) facilitates knowledge transfer from a source domain to a related target domain.
    This paper investigates a practical DA paradigm, namely {S}ource data-{F}ree {A}ctive {D}omain {A}daptation ({SFADA}), where source data becomes inaccessible during adaptation, and a minimum amount of annotation budget is available in the target domain.
    Without referencing the source data, new challenges emerge in identifying the most informative target samples for labeling, establishing cross-domain alignment during adaptation, and ensuring continuous performance improvements through the iterative query-and-adaptation process.
    In response, we present \textit{learn from the learnt (LFTL)}, a novel paradigm for SFADA to leverage the learnt knowledge from the source pretrained model and actively iterated models without extra overhead. 
    We propose \textit{Contrastive Active Sampling} to learn from the hypotheses of the preceding model, thereby querying target samples that are both informative to the current model and persistently challenging throughout active learning. 
    During adaptation, we learn from features of actively selected anchors obtained from previous intermediate models, so that the \textit{Visual Persistence-guided Adaptation} can facilitate feature distribution alignment and active sample exploitation.
    Extensive experiments on three widely-used benchmarks show that our LFTL achieves state-of-the-art performance, superior computational efficiency and continuous improvements as the annotation budget increases.
    Our code is available at https://github.com/lyumengyao/lftl.
\end{abstract}
    
\section{Introduction}
\label{sec:intro}

Deep neural networks have thrived in computer vision but struggle when real-world data deviates from the ideal independent and identical distribution of training data, causing performance drops in well-trained models.
This is where domain adaptation (DA) comes into play, which enables knowledge transfer from a source domain to a target domain of different data distribution. 
Most DA studies assume concurrent access to data from both domains to leverage the inter-domain relationship. Nonetheless, the reliance on labeled source data during adaptation could impede its widespread usage in real-world scenarios, considering stringent data protection regulations and resource deficiency in storage and computation.
Therefore, source-free unsupervised DA (SFUDA)~\cite{liang2021sfuda,li2020sfuda,xia2021sfuda,huang2021sfuda,yang2021model,zhang2022dac,qiu2021CPGA,hwang2024sfda2} advocates for encapsulating knowledge within a source-trained model.
They typically employ parameter sharing and hypothesis-driven learning for model adaptation. Yet, the absence of direct supervision from any domain can intensify the ill-posed nature of unsupervised DA~\cite{shi2012informax}, leading to the performance bottlenecks despite considerable efforts made.

\begin{figure}[t]
    \centering
    \includegraphics[width=\textwidth]{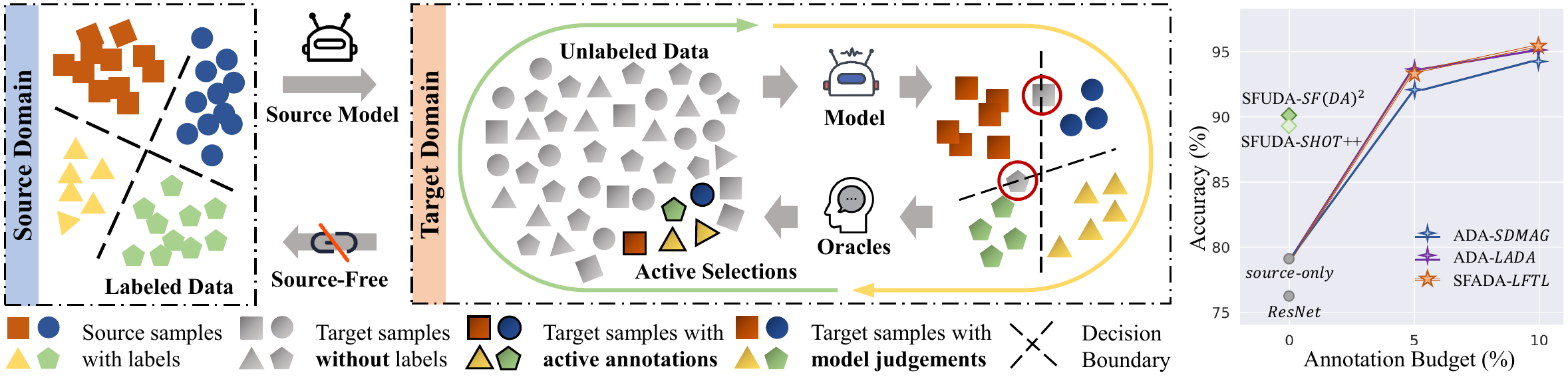}
    \caption{(L) \textbf{S}ource data-\textbf{F}ree \textbf{A}ctive \textbf{D}omain \textbf{A}daptation (SFADA) paradigm. 
    Note that the ratio of labeled/unlabeled targets is much lower ($\le 1\%$ or $\le 5\%$ in our experiments)  than in the illustration. 
    (R) Performance comparison among  DA SOTAs of different settings on Office-31.}
    \label{fig:setting}
\end{figure}

In fact, the presence of the target sample pool during adaptation implies that minimal annotation effort is possible, yielding substantial performance gains.
Hence we study a more practical DA setting, dubbed source-free active DA (SFADA). As shown in Fig~\ref{fig:setting}~(L), the adaptation process is freed from source data, and meantime a minimum amount of annotation budget is available for iteratively querying labels in the target domain. 
Limited yet definitive supervisory signals intrinsically alleviate the ill-posedness phenomenon, and additionally, SFADA offers an appealing trade-off between labeling costs and adaptation performance regarding accuracy and efficiency. 

Taking both target annotation availability and source data inaccessibility into consideration, SFADA presents new challenges.
During query selection, general active learning criteria~\cite{beluch2018ensemble,settles2009activesurveyentropy,sener2018coreset,gal2017bayesian,yan2009kmeans,joshi2009margin,lyu2023box} tend to fail under the domain shift phenomenon as observed in both previous Active DA (ADA) studies~\cite{rai2010ADA,su2020ADA,fu2021ADA,prabhu2021active,Xie2022CVPR,sun2022ADA_lada} and our experiments \ref{sec:exp-comp}. 
On top of that, ADA typically employs source data as a reference to identify distinctive target samples, which violates our source-free setting and thus is not compatible here either. 
Therefore, SFADA necessitates a new active learning strategy to \textit{pinpoint the most innovative and informative target samples utilizing solely a source-trained model.} 
Another challenge comes at the transfer stage. 
Without the source data, manifold alignment, typically employed via minimizing distribution divergence to boost performance, is impeded. Yet the labeling costs have increased the expectations for performance enhancement.
This prompts us to ask \textit{how to exploit the newly acquired knowledge while consolidating the learnt domain-invariant information during adaptation?}
And following these, \textit{how to guarantee continuous performance improvement during the iterative query-and-adaptation process?}

To fill in this gap, this paper formulates a novel SFADA paradigm to \textit{learn from the learnt} (LFTL), as illustrated in Fig.~\ref{fig:pipeline}. We employ the hypothesis and visual representations of target samples obtained from the learnt source model and actively iterated models to ascertain what to supplement and where to adapt in the target domain.
We first conduct \textit{active learning from the learnt model}. 
Although it is impossible to directly identify target samples the most divergent from the source, the hypothesis of learnt model can be utilized as an indication of how well the target domain knowledge is understood. Furthermore, as the model progressively adapts towards the target domain, 
the hypothesis of the current model becomes increasingly important to query newer labels. This has not been addressed by the one-round query of MHPL~\cite{wang2023mhpl}.
Inspired from contrastive decoding~\cite{cd,acd,dola,damonlpsg2023vcd,tad}, we propose \textit{Contrastive Active Sampling (CAS)} to leverage the hypothesis of the model from the previous active round. It emphasizes samples that are both informative to the current model and persistently challenging throughout the iterative process, while deprioritizing samples that are redundant or deliver knowledge already acquired during active selection. Upon the contrastively decoded hypothesis, we take the difference between the best-versus-second best (BvSB) guesses~\cite{joshi2009margin}, integrate a class-balancing factor, so that temporally-stagnant, sample-uncertain and class-minor targets can be effectively queried.
Secondly, we perform \textit{transfer learning from the learnt visual representations}. Since the alignment of cross-domain distributions cannot be directly achieved, we establish \textit{Visual Persistence-guided Adaptation (VPA)} to maintain feature representations of active target samples throughout the whole process, where the understanding initially derived from the source domain and subsequently obtained from previous active rounds are effectively conserved via momentum updates. Then our learning objective encourages unlabeled samples to be centralized around these memory-preserving representatives, approximating a source-similar distribution while exploiting target-specific knowledge obtained from queried labels.

Gaining insights from the intermediate results already computed during iteration makes our method simple in architecture, effective in active learning and knowledge transfer, superior in terms of both accuracy and efficiency, and flexible in the trade-off between annotation plus time budget and adaptation accuracy.
Take the performance on the VisDA-C~\cite{peng2017visdac} benchmark for example, the entire query-and-adaptation process obtains 87.4\% with merely 1\% labeling budget and 780 training iterations (0.3h). In contrast, SHOT++ operating under the SFUDA setting requires 21.6K iterations (5.8h) to achieve 87.3\%. Even when factoring in the estimated annotation time (1.83h), our method is significantly more time-saving. 
Compared with LADA~\cite{sun2022ADA_lada}, the ADA state-of-the-art (SOTA), we achieve comparable performance without accessing the source data, while benefiting from a 25\% deduction in active sampling time and an approximately 17-fold increase in adaptation speed. Furthermore, it exceeds prior SFADA work MHPL~\cite{wang2023mhpl} by a clear margin of 1.5\%.
Our main contributions are: 
\begin{itemize}
    \item a novel SFADA paradigm to learn from the learnt source model and actively iterated models, which frees it from specialized architecture and sophisticated learning schemes with superior accuracy and efficiency;
    \item a CAS strategy to learn from previous model hypothesis, thereby prioritizing targets that are confusing for the current model, less transferable in class membership and consistently challenging in previous active rounds;
    \item a VPA design to learn from previous feature representations, so that intrinsic distribution alignment and active sample exploitation can be achieved;
    \item SOTA adaptation accuracy, superior computational efficiency, and continual improvements extensively validated on varied-scaled benchmarks.
\end{itemize}

\section{Related Work}
\label{sec:relatedwork}

\textbf{Domain Adaptation (DA)} is a specialized case of transfer learning that enables knowledge generalization from a source domain to a related target domain. Varying in the assumption of the label-set relationship between source and target domains, namely semantic shift, DA settings can be mainly divided into close-set DA, open-set DA~\cite{panareda2017ODA,saito2018ODA}, partial DA~\cite{cao2018PDA,li2020PDA} and universal DA~\cite{you2019UDA,saito2020UDA}. According to the assumption of target data accessibility, DA settings fall into unsupervised DA (UDA)~\cite{ganin2016uda,xiong2023confidence}, semi-supervised DA (SSDA)~\cite{saito2019ssda,he2020ssda}, weakly-supervised DA (WSDA)~\cite{tan2019wda}, zero-shot DA~\cite{peng2018zda,wang2019zda}, one-shot DA~\cite{luo2020oda}, few-shot DA~\cite{xu2019fda,motiian2017fda,hao2024qprompt} as well as active DA (ADA)~\cite{rai2010ADA,su2020ADA,fu2021ADA,prabhu2021active,Xie2022CVPR,sun2022ADA_lada}. 

Among the subfields, the one most related to our setting is \textbf{ADA}, where the goal is to query the most informative target samples for annotation to best benefit classification on the target domain. 
Saha \etal~\cite{saha2011ADA} either infer pseudo labels on target samples or query them for oracle annotation, which is based on inter-domain similarity between source and target domains. Su \etal \cite{su2020ADA} train a domain discriminative model for domain alignment and target uncertainty estimation. While most of them utilize or are orthogonal to off-the-shelf active learning mechanisms, Fu \etal~\cite{fu2021ADA} argue that constructing a committee~\cite{melville2004qbc}, \ie ensemble to measure target sample uncertainty based on consensus between multiple predictions is more robust under domain shift.
CLUE~\cite{prabhu2021active} is a k-means clustering-weighted uncertainty-based  active learning strategy, and it optimizes with cross-domain minmax entropy~\cite{saito2019ssda}.
Xie \etal~\cite{Xie2022CVPR} apply the margin loss to the source domain training to exploit hard source samples and a less domain-biased decision boundary, and they further supplement margin sampling with expected error reduction consistent with the training objective.
LADA~\cite{sun2022ADA_lada} bases its sampling criterion on the predictive purity of a local structure, and exploits both data sources during adaptation.
Despite previous efforts, ADA methods typically necessitate the simultaneous presence of data from both the source and target domains to exploit the relationship between domains. However, when this assumption fails in practical scenarios, we must tackle the challenges associated with the unavailability of source data.

On the other hand, \textbf{SFUDA} posits a transfer setting where neither the source data nor the target annotations are available.
Kundu \etal~\cite{kundu2020sfunida} adopt a two-stage paradigm for universal DA, where the procurement stage trains a source model in consideration of future semantic shifts and domain gaps, so that the deployment stage is capable of operating without source data. 
Liang \etal~\cite{liang2021sfuda} harness information maximization~\cite{shi2012informax,hu2017IM} to align different domains, and utilize self-supervision and curriculum learning techniques via pseudo label clustering.
Xia \etal~\cite{xia2021sfuda} introduce adversarial training  to distinguish between source-similar and -dissimilar target samples for cross-domain alignment, facilitated with self-supervision and clustering to improve performance.
Based on prediction confidence, DaC~\cite{zhang2022dac} also divides the target samples based on source similarity, employing a global structure for source-like targets while implementing a local structure for the distinctive samples.
Without definitive supervision signal from either domain, SFUDA makes the alignment more challenging and exacerbates the ill-posedness of UDA. 
Consequently, it frequently demands intricate design and increased computations to attain desirable outcomes.

Given the accessibility of target samples, this paper focuses on the more practical \textbf{SFADA} paradigm, leveraging minimal resources for better performance. 
Prior work MHPL~\cite{wang2023mhpl} integrates three active learning strategies to sample targets that are uncertain, diverse and different from the source domain, and then utilizes a neighbor focal loss to emphasize them in adaptation. However, the dependency on the source dissimilarity metric confines it to a one-round sampling method, failing to guarantee sustained performance improvement in real-world applications.
SALAD~\cite{kothandaraman2023salad} samples target with a binary weighted function between an entropy score and the expected model change. However the adaptation procedure incorporates an additional Guided Attention Transfer Network, leading to undesirable computational overhead and, as evidenced by the results, inefficient annotation utilization.
In contrast, our LFTL offers a straightforward, broadly applicable, and growth-oriented solution that can make the best from previous knowledge and a limited budget to guarantee accuracy, efficiency, and a flexible trade-off between annotation resources and performance.

\section{Learn from the Learnt}
\label{sec:method}

\begin{figure}[t]
\centering
\includegraphics[width=0.9\linewidth]{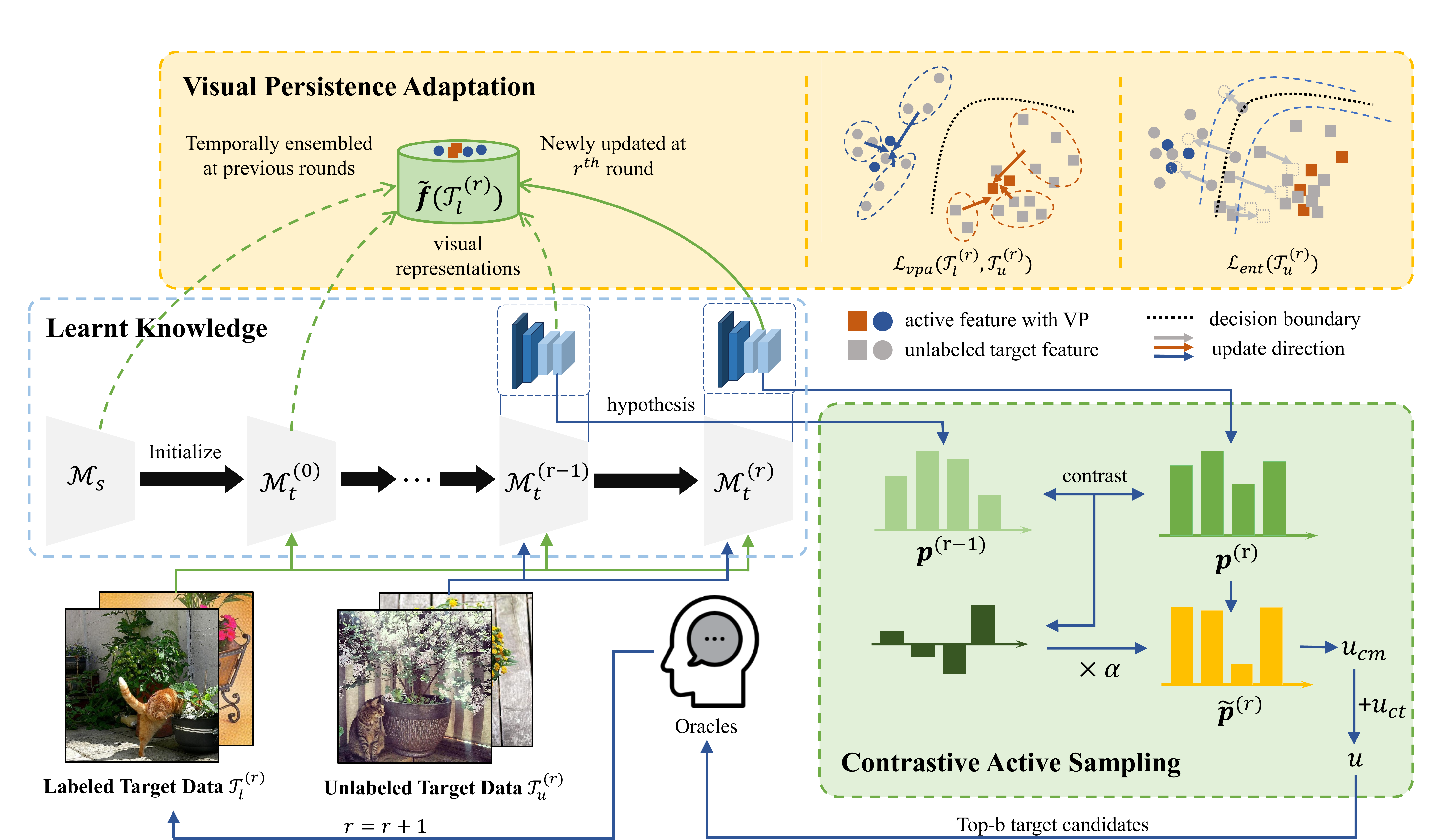}
\caption{The proposed LFTL framework for SFADA. Contrastive Active Sampling emphasizes freshly acquired knowledge in the posterior distribution so that novel samples are more likely to be queried for annotations. Then during adaptation the persistence vault retains previous domain-invariant knowledge to facilitate alignment in the target domain via $\mathcal{L}_{vpa}$ and $\mathcal{L}_{ent}$. As the iterative process continues, it yields more informative targets and an improved target model.}
\label{fig:pipeline}
\end{figure}

In this section, we present \textit{LFTL} for SFADA. As illustrated in Fig.~\ref{fig:pipeline}, it alternates between Contrastive Active Sampling (CAS)  and Visual Persistence-guided Adaptation (VPA). 

\subsection{Overview}

Given a labeled source domain $\mathcal{S}$ with restricted access  and an unlabeled target domain $\mathcal{T}$, our goal is to derive a model that can minimize the risk on the target domain through knowledge transfer with a small annotation budget.
At the source end, image and label pairs $\mathcal{S}=\{(x_s, y_s)\}$ from the source domain are first utilized to train a model $\mathcal{M}_s: x_s\mapsto y_s$, 
which consists of a feature extractor $f(\cdot)_s$ and a classifier $g(\cdot)_s$, \ie $\mathcal{M}_s(x_s) = g_s(f_s(x_s))$.
After severing communication with the source domain data, we adapt $\mathcal{M}_s$, where the prior knowledge is maintained, to the target domain with the assistance of a budget of $B$ labeled instances. The initial target dataset has a total of $n_{tu}^{(0)}$ unlabeled samples $\mathcal{T}_u^{(0)}=\{x_{tu}^i\}_{i=1}^{i=n_{tu}^{(0)}}$ and a shared label space $\mathcal{Y_T}$.

At the beginning of the $R$-round query-and-adaptation alternation ($r=0$), the target model $\mathcal{M}_{t}^{(r)}$, without any architectural adaptation, is initialized from $\mathcal{M}_{s}$ as $\mathcal{M}_{t}^{(0)}$.
We actively query labels for $b=B/R$ most informative target instances with the criterion of CAS (Sec.~\ref{sec:method-active}), which initializes the labeled subset of the target pool, denoted as $\mathcal{T}_l^{(1)}={\{x_{tl}^j\}_{j=1}^{j=n_{tl}^{(1)}}}$, and thus $\mathcal{T}_u^{(1)}=\mathcal{T}_u^{(0)}\setminus\mathcal{T}_l^{(1)}$. Then the parameters of the target model can be optimized using the newly acquired target data splits, and updated to $\mathcal{M}_{t}^{(1)}$ via the proposed VPA strategy (Sec.~\ref{sec:method-opt}). 
Likewise, in subsequent rounds ($r\geq1$), target data splits $\mathcal{T}_l^{(r)} \leftarrow \mathcal{T}_l^{(r-1)}\cup \{x_{tl}^j\}_{n_{tl}^{(r)}}$ and $\mathcal{T}_u^{(r)}=\mathcal{T}_u^{(r-1)}\setminus\mathcal{T}_l^{(r)}$, as well as the task model $\mathcal{M}_{t}^{(r)}$ are updated, continuously querying novel knowledge and optimizing towards the target domain through the synergistic interaction of sampling and adaptation.

\subsection{Contrastive Active Sampling}
\label{sec:method-active}
In each active learning round, the aim is to find the most informative instances from the target domain, especially those diverging from the source, to benefit the adaptation of the model. 
Without directly referencing the source data,  previous source-free approaches~\cite{xia2021sfuda,liang2021sfuda,li2020sfuda,Ding_sfuda_distestimate} often rely on the hypothesis transfer to distinguish between source-like and source-dissimilar samples. 
However, when applied to the SFADA scenario, such solution~\cite{wang2023mhpl} is constrained to a single round of active sampling and model updating, which restricts the potential applications of this setting. 

Considering the iterative nature of the task context, we tackle this challenge from the active perspective. Just as the source knowledge is encapsulated within the initial model $\mathcal{M}_s$, the target knowledge learned in previous rounds is manifested in the hypotheses of the preceding model $\mathcal{M}_{t}^{(r-1)}$. 
Compared to the hypotheses of the preceding model, samples that obtain higher prediction confidence from the newly updated model reflect the fresh insight just acquired. 
The larger the gap in predictive confidence,
the better the current model has grasped the sample, and thus the less informative for  the subsequent phase of sample selection.
As illustrated in Fig~\ref{fig:pipeline}, we first contrastively highlight the less informative samples as follows:
\begin{equation}
    \mathbf{\Tilde{p}}^{(r)}(\cdot|x_{tu}^i) = 
    \begin{cases} 
        \log \mathbf{p}^{(r)} & \text{if } r = 0, \\
        \log \mathbf{p}^{(r)} + \alpha (\log \mathbf{p}^{(r)} - \log \mathbf{p}^{(r-1)}) & \text{if } r > 0,
    \end{cases}
    \label{eq:contrastive}
\end{equation}
weighted by $\alpha$, where 
$\mathbf{p}$ is derived from the softmax of the model logits 
$\delta(g(f(x_{tu}^i)))$. 
We omit the superscript of round $(r)$ for simplicity in the following text.

Based on $\mathbf{\Tilde{p}}$, various active learning criteria can be employed. Without loss of generality, here we focus on the most confused classes. Specifically, we take the difference between the best-versus-second best (BvSB) guesses as the uncertainty indicator, defined as $u_{cm}(x_{tu}^i)=\Tilde{p}(y_a^i|x_{tu}^i) - \Tilde{p}(y_b^i|x_{tu}^i)$, where $y_a^i = \arg\max_{y\in \mathcal{Y}} \Tilde{p}(y|x_{tu}^i)$, and $
    y_b^i = \arg\max_{y\in \mathcal{Y}\setminus y_a^i} \Tilde{p}(y|x_{tu}^i)$.
Then the unlabeled target sample pool is ranked by:
\begin{equation}
    \mathbf{r}(x_{tu}^i) = \sum_{k=1}^{n_{tu}} \mathbbm{I}(u_{cm}(x_{tu}^k) < u_{cm}(x_{tu}^i)),
    \label{eq:rank1}
\end{equation}
where $\mathbbm{I}$ is the indicator function.
A smaller $u_{cm}$ score indicates that the assessed target sample $x_{tu}^i$ furnishes novel insights that diverge from the source domain and, in contrast to prior iterations, has not been grasped. Conversely, when the current stronger model, aided by a marginal increment of active labels, assigns more probability preference to one class for an unlabeled sample, CAS emphasizes it via a larger $u_{cm}$ score, lowering the priority of those familiar candidates.
As the query-and-adaptation alternation proceeds, CAS continuously harvests fresh information for transfer by leveraging readily available intermediate results, without additional extra computational burden or memory usage.

The contrastively decoded margin
evaluates each sample on an individual basis. 
However, the batch-mode active learning could be susceptible to outliers, since the queried samples are not selected based on the underlying natural density distribution~\cite{baum1992query,settles09AL}.
Particularly in DA tasks, different classes show different sensitivity to the domain shift. Some of them are more robust to certain shifts, \ie their learnt features present more domain-invariant patterns, whereas others are likely to suffer a steep performance drop~\cite{jin2020VDA}. 
These motivate us to incorporate a broader, semantic view of the target domain into our active criterion, where classes with higher transferability is estimated via the class memberships of reliable model hypotheses:
\begin{equation}
    u_{ct}(c|\mathcal{T}_u) = \frac{\sum_{i=1}^{n_{tu}}\mathbbm{I}(y^i_a=c)\cdot \mathbbm{I}(\mathbf{r}(x_{tu}^i)>(n_{tu}-\kappa))}
    {\max_c[\sum_{i=1}^{n_{tu}}\mathbbm{I}(y^i_a=c)\cdot \mathbbm{I}(\mathbf{r}(x_{tu}^i)>(n_{tu}-\kappa))]}.
    \label{eq:stat}
\end{equation}
Eq.~\ref{eq:stat} measures the frequency of unlabeled samples being inferred as class $c$ among $\kappa$ highest contrastive margins, wherein $\kappa$ controls the range of statistics. Thus the classes with higher $u_{ct}$ scores can be downweighted in favor of less transferable classes.

In this way, we drive a hybrid active sampling criterion that favors target samples with local uncertainty, global intransferability and temporal stagnancy:
\begin{equation}
    \label{eq:total_sample}
    u^i = u_{cm}^i + \lambda u_{ct}({y^i_a}),
\end{equation}
where $\lambda$ controls the importance of class-level cross-domain transferability.

\subsection{Visual Persistence-guided Adaptation}
\label{sec:method-opt}

During each iterative round, the top-$b$ most informative samples are queried for labels towards the adaption to the target domain. They offer reliable signals but also pose a challenge of making the best use of them. 
We first warm up the adaptation process via empirical risk minimization on the small amount of target instances with active labels:
\begin{equation}
    \label{eq:ce}
    \mathcal{L}_{ce} = -\mathbbm{E}_{x_{tl}\sim\mathcal{T}_l}\: \mathbf{q}(x_{tl})^T\log[\mathbf{p}(\cdot|x_{tl})],
\end{equation}
where $\mathbf{q}(x_{tl})$ is the ground truth 1-hot vector for the labeled instance $x_{tl}$.

To further minimize the risk on the target domain,
we are motivated to bridge the distance between actively queried data and unlabeled data.
Instead of seeking cluster centroids in the target pool, we directly take active samples as representative anchors, and encourage unlabeled data to form concentrated clusters around the nearest anchor on the visual embedding space via soft similarity minimization:
\begin{gather}
    \label{eq:fs}
    \mathcal{L}_{ac} = -\mathbbm{E}_{x_{tu}\sim\mathcal{T}_u}\: \mathbf{d}_{tu}^T\log(\mathbf{d}_{tu}),
    \text{where}\;\; \mathbf{d}_{tu} = \delta[\mathcal{D}(f_e(x_{tu}), \mathbf{f}(\mathcal{T}_l))].
\end{gather}
In Eq.~\ref{eq:fs}, the feature extractor of the adapted model outputs the visual representation of 
$x_{tu}$ as $f_t(x_{tu})$. $\mathbf{f}(\mathcal{T}_l)$ is a $n_{tl}\times d$ matrix where each row represents a $d-$dimensional feature of a labeled anchor $x^i_{tl}$. $\mathcal{D}$ denotes a distance metric. For simplicity and without loss of generality, we adopt cosine similarity to measure the relationship between different instances, and a normalization function is applied to transform the distances to $[0,1]$. 
In decreasing the entropy of distances, each unlabeled sample is encouraged to get close to its nearest labeled anchors and meantime keep other clusters at a considerable distance. Consequently, as shown in Fig.~\ref{fig:pipeline}, features are aggregated in a class-wise manner, which makes it easier for classifier decision boundaries to be established.

However, in the context of the SFADA framework, as the labeled target subset $\mathcal{T}_l$ and the model $\mathcal{M}_t$ iterate for $R$ rounds, the incremental domain-specific information obtained from the actively queried samples may intensify the cross-domain incompatibility~\cite{tzeng2017adda,shu2018dirt,chang2019dsbn}. Furthermore, there is a potential risk of domain-invariant knowledge being forgotten during the process.
Without accessing the source domain data, previous source-free methods achieve domain alignment via freezed hypotheses~\cite{xia2021sfuda,liang2021sfuda,li2020sfuda} or buffered embeddings~\cite{Ding_sfuda_distestimate} from the source classifier. In our multi-round iterative process, we introduce a visual persistence vault to guide the adaptation. It preserves judgements on the active anchors, which have been derived from source and intermediate models, via exponential moving average: 
\begin{equation}
    \mathbf{\tilde{f}}(x_{tl}) \leftarrow \gamma \mathbf{f}(x_{tl}) + (1-\gamma)\mathbf{\tilde{f}}(x_{tl}), 
\end{equation}
where $\mathbf{f}(x_{tl})$ is derived from the adapted feature extractor $f_t$, and $\gamma$ is the momentum parameter controlling the weighting decrease of previous observations, which is empirically fixed as $0.9$. 
In replacing the most recent feature representation of active anchors $\mathbf{f}(\mathcal{T}_l)$ in Eq.~\ref{eq:fs} with the temporally ensembled $\tilde{\mathbf{f}}(\mathcal{T}_l)$, we obtain the visual persistence-guided adaption loss $\mathcal{L}_{vpa}$. The learnt domain-invariant knowledge maintained by the persistence vault effectively supports the alignment in the target domain, and meantime the target-specific information is well exploited.

Additionally, the entropy minimization loss~\cite{liang2021sfuda,wang2023mhpl,huang2021sfuda} is introduced to approach the ideal adaptation performance and foster discriminative features in the learnt manifold:
\begin{equation}
    \label{eq:ls}
    \mathcal{L}_{ent} = -\mathbbm{E}_{x_{tu}\sim\mathcal{T}_u}\: \mathbf{p}(\cdot|x_{tu})^T\log[\mathbf{p}(\cdot|x_{tu})].
\end{equation}

Overall, the combination of supervised cross-entropy, visual persistence guidance and entropy minimization losses yields:
\begin{equation}
    \label{eq:all}
    \mathcal{L} = \mathcal{L}_{ce} + \beta_1\mathcal{L}_{vpa} + \beta_2\mathcal{L}_{ent}.
\end{equation}
We emphasize that no explicit supervision (\eg pseudo labels) is enforced for unlabeled samples during the optimization, so that the error accumulation problem is greatly alleviated in the source data-free and target label-scarce setting.

\section{Experiments}
\label{sec:exp}
\begin{table}[tb]
\caption{Results on the large-scale VisDA-C dataset (ResNet101) in terms of classification accuracy ($\%$). SF represents source inavailability, and AS shows percentage of active annotations ($\%$). Best results are highlighted in bold, and the best in each section are underlined.}
\label{tab:visdac}
\centering
\small
\setlength{\tabcolsep}{1.0mm}{
\begin{adjustbox}{max width=1.0\textwidth}
  \newcolumntype{g}{>{\columncolor{mygray}}c}
  \begin{tabular}{lccccccccccccccg}
    \toprule
    Method & SF & AS & plane &bcycl &bus &car &horse &knife &mcycl &person &plant &sktbrd &train &truck &Per-class \\
    \midrule
        SFDA~\cite{kim2020sfda} &\checkmark& - & 86.9 &81.7 &84.6 &63.9 &93.1 &91.4 &86.6 &71.9 &84.5 &58.2 &74.5 &42.7& 76.7 \\
        A$^2$Net~\cite{xia2021sfuda}&\checkmark& - & 94.0 &87.8 &85.6 &66.8 &93.7 &95.1 &85.8 &81.2 &91.6 &88.2 &86.5 &56.0& 84.3 \\
        SHOT~\cite{liang2021sfuda} &\checkmark& - & 95.8 &88.2 &87.2 &73.7 &95.2 &96.4 &87.9 & \underline{84.5} & 92.5 &89.3 &85.7 &49.1 &85.5 \\
        SHOT$++$~\cite{liang2021sfuda} &\checkmark& - & \underline{97.7} &{88.4} &\underline{90.2} &\underline{86.3} &\underline{97.9} &\underline{\bf{98.6}} &{92.9} & 84.1 & \underline{97.1} & {92.2} &\underline{93.6} &28.8 &{87.3} \\
        CPGA~\cite{qiu2021CPGA} &\checkmark& - & 94.8 &83.6 &79.7 &65.1 &92.5 &94.7 &90.1 &82.4 &88.8 &88.0 &88.9 &{60.1} &84.1 \\
        DaC~\cite{zhang2022dac} &\checkmark& - & 96.6 &86.8& 86.4& 78.4& 96.4& 96.2& \underline{93.6}& 83.8& 96.8& \underline{95.1}& 89.6& 50.0& 87.3 \\
        SF(DA)$^2$~\cite{hwang2024sfda2} &\checkmark& - & 96.8&  \underline{89.3}&  82.9&  81.4&  96.8&  95.7&  90.4&  81.3&  95.5&  93.7&  88.5&  \underline{64.7}&  \underline{88.1} \\
    \midrule
        AADA~\cite{su2020ADA} &\tiny\XSolidBrush& 1 & 83.5 & 64.0 & 67.2 & 80.5 & 87.8 & 61.4 & 88.5 & 79.1 & 87.9 & 78.5 & 84.7 & 32.6 & 74.6\\
        TQS~\cite{fu2021ADA} &\tiny\XSolidBrush& 1 & 87.3 & 77.5 & 77.1 & 67.0 & 90.5 & 58.4 & 81.1 & 82.0 & 91.1 & 73.9 & 68.7 & \underline{57.6}  & 76.0\\
        CLUE~\cite{prabhu2021active} &\tiny\XSolidBrush& 1 & 77.0 & 57.8 & 73.4 & 76.5 & 76.9 & 68.7 & 87.0 & 75.5 & 85.1 & 66.5 & 76.1 & 47.3 & 72.3\\
        SDM-AG~\cite{Xie2022CVPR} &\tiny\XSolidBrush& 1 & 84.0 & 72.3 & 77.8 & 82.0 & 92.2 & 81.5 & 83.1 & 71.4 & 85.2 & 74.2 & 80.2 & 35.1 & 76.6\\
        LADA~\cite{sun2022ADA_lada} &\tiny\XSolidBrush& 1 & \underline{97.8} & \underline{82.3} & \underline{92.0} & \underline{86.8} & \underline{98.0} & \underline{94.9} & \underline{94.7} & \underline{88.8} & \underline{95.6} & \underline{95.0} & \underline{93.9} & 57.0 & \underline{89.7}\\
    \cdashlinelr{1-16}
        AADA~\cite{su2020ADA} & \tiny\XSolidBrush & 5 & 92.3 & 78.5 & 87.4 & 87.4 & 91.8 & 90.8 & 91.2 & 86.2 & 92.6 & 90.3 & 90.1 & 61.5 & 86.7\\
        TQS~\cite{fu2021ADA} &\tiny\XSolidBrush& 5 &  89.6 & 85.8 & 82.9 & 78.5 & 96.8 & 82.8 & 90.2 & 81.6 & 93.9 & 85.2 & 87.0 & \underline{69.6} & 85.3\\
        CLUE~\cite{prabhu2021active} &\tiny\XSolidBrush& 5 & 92.8 & 81.7 & 83.0 & 84.0 & 93.4 & 89.1 & 91.4 & 89.1 & 94.2 & 88.0 & 85.3 & 62.0 & 86.2\\
        SDM-AG~\cite{Xie2022CVPR} &\tiny\XSolidBrush& 5 & 93.3 & 85.1 & 83.7 & 86.7 & 94.2 & 90.2 & 92.8 & 85.5 & 90.2 & 81.1 & 80.8 & 56.2 & 85.0\\
        LADA~\cite{sun2022ADA_lada} &\tiny\XSolidBrush& 5 & \textbf{\underline{98.6}} & \underline{87.4} & \underline{\bf{92.9}} & \underline{\bf{91.3}} & \underline{\bf{98.6}} & \underline{97.0} & \underline{\bf{96.0}} & \underline{91.8} & \underline{97.4} & \underline{\bf{97.2}} & \underline{\bf{95.2}} & 61.6 & \underline{92.1}\\
    \midrule
        LFTL & \checkmark& 1 & 95.9 & 84.6 & 84.6 & 77.1 & 95.4 & 93.6 & 91.4 & 87.1 & 93.2 & 90.4 & 87.8 & 67.6 & 87.4 \\
    \cdashlinelr{1-16}
        MHPL~\cite{wang2023mhpl} & \checkmark& 5 & - & - & - & - & - & - & - & - & - & - & - & - & 91.3 \\
        LFTL  & \checkmark& 5 & \underline{98.0} & \underline{\bf{92.5}} & \underline{88.7} & \underline{89.1} & \underline{98.0} & \underline{97.2} & \underline{94.3} & \underline{\bf{93.5}} & \underline{\bf{98.0}} & \underline{96.5} & \underline{92.6} & \underline{\bf{75.6}} & \underline{\bf{92.8}} \\
    \bottomrule
  \end{tabular}
\end{adjustbox}
}
\end{table}

\subsection{Setup}
\myparagraph{Datasets.}
We conduct experiments on three widely-used DA benchmarks: 
a) \textbf{VisDA-C}~\cite{peng2017visdac} aims to solve the simulation-to-reality shift, in which the source domain contains 152K synthetic images with varying angles and lightning conditions, and the target domain is composed of 55K images cropped from the Microsoft
COCO dataset~\cite{lin2014microsoftcoco}, each with the same 12 categories.
b) \textbf{Office-Home}~\cite{venkateswara2017officehome} is a medium-sized dataset consisting of 4 domains: Artistic (Ar), Clip Art (Cl), Product (Pr) and Real-World (Rw) images, each with the same 65 classes.
c) \textbf{Office-31}~\cite{saenko2010office31} is small-sized, which consists of 31 classes in 3 different domains: Amazon (A), DSLR (D) and webcam (W).

\myparagraph{Active Setting.} Taking the dataset scales into consideration, we budget 1\% and 5\% of VisDA-C for annotation. For the medium- and small-scaled office datasets, 5\% and 10\% data is actively sampled. In each round, 0.1\%  annotation increment is made.
In all experiments, the average performance of 3 repeated trials is reported.

\myparagraph{Implementation.}
Following SHOT~\cite{liang2021sfuda}, A$^2$Net~\cite{xia2021sfuda}, CPGA~\cite{qiu2021CPGA} and MHPL~\cite{wang2023mhpl}, our main results are reported on ResNet101~\cite{he2016resnet} for VisDA-C and ResNet50 for Office, on top of which is a bottleneck layer that compresses the 2048 high dimensional features into 256 units, followed by a task-dependent FC layer.
In the source domain, we randomly split source data into 90\% training and 10\% validation to obtain a well-trained model. 
Considering that the number of active selections is quite small, we harness Mixup~\cite{zhang2018mixup} regularization to exploit the potential of data during target adaptation.
Component verification and more training details can be found in supplementary.

\begin{table}[tb]
\centering
\caption{Results on the Office-Home dataset (ResNet50) in terms of classification accuracy ($\%$). SF represents source inavailability, and AS shows percentage of active annotations ($\%$). Best results are highlighted in bold, and the best in each section are underlined.}
\label{tab:officehome}
\setlength{\tabcolsep}{0.6mm}{
\begin{adjustbox}{max width=1.0\textwidth}
  \newcolumntype{g}{>{\columncolor{mygray}}c}
  \begin{tabular}{lccccccccccccccg}
    \toprule
    Method & SF & AS &Ar$\to$Cl & Ar$\to$Pr & Ar$\to$Rw & Cl$\to$Ar & Cl$\to$Pr &Cl$\to$Rw & Pr$\to$Ar &Pr$\to$Cl &Pr$\to$Rw &Rw$\to$Ar &Rw$\to$Cl &Rw$\to$Pr & Avg \\
    \midrule
        SFDA~\cite{kim2020sfda} &\checkmark& - &48.4 &73.4 &76.9 &64.3 &69.8 &71.7 &62.7 &45.3 &76.6 &69.8 &50.5 &79.0 &65.7 \\ 
        A$^2$Net~\cite{xia2021sfuda}&\checkmark& -&58.4 &79.0 &82.4 &67.5 &79.3 &78.9 &68.0 &56.2 &82.9 &\underline{74.1} &60.5 &\underline{85.0} &72.8 \\
    	SHOT~\cite{liang2021sfuda} &\checkmark& -&57.7 &79.1 &81.5 &67.6 &77.9 &77.8 &68.1 &55.8 &82.0 &72.8 &59.7 &84.4 &72.0 \\
    	SHOT++~\cite{liang2021sfuda} &\checkmark& -&{57.9} &\underline{79.7} &\underline{82.5} &68.5 &\underline{79.6} &\underline{79.3} &\underline{68.5} &{57.0} &\underline{83.0} &73.7 &{60.7} &84.9 &\underline{73.0} \\
    	CPGA~\cite{qiu2021CPGA} &\checkmark& - & \underline{59.3} &78.1 &79.8 &65.4 &75.5 &76.4 &65.7 &\underline{58.0} &81.0 &72.0 &\underline{64.4} &83.3 &71.6 \\
        DaC~\cite{zhang2022dac} &\checkmark& - & 59.1 & 79.5 & 81.2 & \underline{69.3} & 78.9 & 79.2 & 67.4 & 56.4 & 82.4 & 74.0 & 61.4 & 84.4 & 72.8 \\
    \midrule
        AADA~\cite{su2020ADA} &\tiny\XSolidBrush& 5 &56.6 &78.1 &79.0 &58.5 &73.7 &71.0 &60.1 &53.1 &77.0 &70.6 &57.0 &84.5 &68.3\\
        TQS~\cite{fu2021ADA} &\tiny\XSolidBrush& 5 & 58.6 &81.1 &81.5 &61.1 &76.1 &73.3 &61.2 &54.7 &79.7 &73.4 &58.9 &86.1 &70.5 \\
        CLUE~\cite{prabhu2021active} &\tiny\XSolidBrush& 5 & 50.7 & 76.1 & 78.1 & 62.6 & 75.2 & 71.0 & 64.0 & 52.3 & 79.7 & 72.7 & 56.9 & 83.8 & 68.6\\
        SDM-AG~\cite{Xie2022CVPR} &\tiny\XSolidBrush& 5 & 61.2 & 82.2 & 82.7 & 66.1 & 77.9 & 76.1 & 66.1 & 58.4 & 81.0 & 76.0 & 62.5 & 87.0 & 73.1\\
        LADA~\cite{sun2022ADA_lada} &\tiny\XSolidBrush& 5 & \underline{71.2} & \underline{87.4} & \underline{84.6} & \underline{72.1} & \underline{87.0} & \underline{83.6} & \underline{71.5} & \underline{71.6} & \underline{85.3} & \underline{79.3} & \underline{75.5} & \underline{90.4} & \underline{80.0} \\
        \cdashlinelr{1-16}
        AADA~\cite{su2020ADA} & \tiny\XSolidBrush & 10 &65.8 &84.5 &82.2 &64.1 &80.6 &76.1 &67.6 &62.6 &80.1 &73.7 &66.1 &88.6 &74.3 \\
        TQS~\cite{fu2021ADA} &\tiny\XSolidBrush& 10 &68.0 &87.7 &85.7 &67.0 &83.0 &78.7 &69.3 &64.5 &83.9 &77.8 &68.9 &90.6 &77.1 \\
        CLUE~\cite{prabhu2021active} &\tiny\XSolidBrush& 10 & 62.1 & 79.1 & 80.3 & 64.1 & 77.1 & 76.7 & 65.4 & 63.9 & 84.2 & 74.2 & 68.6 & 84.7 & 73.4\\
        SDM-AG~\cite{Xie2022CVPR} &\tiny\XSolidBrush& 10 & 68.5 & 87.6 & 86.4 & 69.5 & 84.8 & 81.2 & 71.0 & 66.2 & 84.8 & 79.3 & 69.9 & 90.8 & 78.3 \\
        LADA~\cite{sun2022ADA_lada} &\tiny\XSolidBrush& 10 & \underline{\textbf{77.2}} & \underline{91.9} &  \underline{88.1} & \underline{76.9} & \underline{91.1} &  \underline{86.8} & \underline{76.6} & \underline{\bf{78.1}} &  \underline{88.3} &  \underline{82.0} & \underline{\bf{79.0}} &  \underline{93.8} & \underline{84.2} \\
    \midrule
        MHPL~\cite{wang2023mhpl} & \checkmark& 5 & \underline{69.0} &  85.7 &  \underline{86.4} &  72.6 &  \underline{87.4} &  84.2 &  \underline{73.3} &  \underline{67.4} &  \underline{86.4} &  \underline{80.1} &  \underline{69.6} &  89.8 &  \underline{79.3}\\
        LFTL & \checkmark& 5 & 66.9 & \underline{86.6} & 85.5 & \underline{73.1} & 86.3 & \underline{84.5} & 72.2 & 65.7  &85.9  &79.2  &69.0  &\underline{90.2} & 78.8 \\
        \cdashlinelr{1-16}
        LFTL & \checkmark& 10 & {76.6} & {\bf{92.2}} & {\bf{89.7}} & {\bf{78.9}} & {\bf{93.0}} & {\bf{89.2}} & {\bf{78.6}} & {77.1} & {\bf{90.0}} & {\bf{83.4}} & {77.8} & {\bf{94.6}} & {\bf{85.1}}\\
    \bottomrule
  \end{tabular}
\end{adjustbox}
}
\end{table}

\begin{figure*}[!t] %
  \begin{minipage}[]{0.54\linewidth}
  
\captionof{table}{Results on the Office-31 dataset (ResNet50) in terms of classification accuracy (\%). SF represents source inavailability, and AS shows percentage of active annotations (\%). Best results are highlighted in bold, and the best in each section are underlined.}
\label{tab:office31}
\begin{adjustbox}{max width=1\textwidth}
  \newcolumntype{g}{>{\columncolor{mygray}}c}
  \begin{tabular}{lccccccccg}
\toprule
    Method & SF & AS &A$\to$D & A$\to$W & D$\to$A & D$\to$W & W$\to$A &W$\to$D & Avg \\
\midrule
    SFDA~\cite{kim2020sfda} &\checkmark& - &92.2 &91.1 &71.0 &98.2 &71.2 &99.5 &87.2\\
    A$^2$Net~\cite{xia2021sfuda}&\checkmark& - &94.5 &94.0 &\underline{76.7} &\underline{99.2} &76.1 &\bf\underline{100.0}&\underline{90.1}\\
    SHOT~\cite{liang2021sfuda} &\checkmark& - &93.9 &90.1 &75.3 &98.7 &75.0 &99.9 &88.8 \\
    SHOT++~\cite{liang2021sfuda} &\checkmark& - &94.3 &90.4 &76.2 &98.7 &75.8 &99.9 &89.2 \\
    CPGA~\cite{qiu2021CPGA} &\checkmark& - &94.4 &\underline{94.1} &76.0 &98.4 &{76.6} &99.8 &89.9 \\
    SF(DA)$^2$~\cite{hwang2024sfda2} &\checkmark& - & \underline{95.8} & 92.1 & 75.7 & 99.0 & \underline{76.8} & 99.8 & 89.9\\
\midrule
    AADA~\cite{su2020ADA} &\tiny\XSolidBrush& 5 & 89.2 &87.3 &78.2 &99.5 &78.7 &\bf{\underline{100.0}} &88.8 \\
    TQS~\cite{fu2021ADA} &\tiny\XSolidBrush& 5 & 92.8 &92.2 &80.6 &\bf{\underline{100.0}} &80.4 &\bf{\underline{100.0}} &91.1\\
    CLUE~\cite{prabhu2021active} &\tiny\XSolidBrush& 5 & 90.7 & 94.3 & 78.7 & 99.1 & 76.1 & 99.8 & 89.8 \\
    SDM-AG~\cite{Xie2022CVPR} &\tiny\XSolidBrush& 5 &  93.5 &94.8 &81.9 & \bf{\underline{100.0}} &81.9 & \bf{\underline{100.0}} & 92.0\\
    LADA~\cite{sun2022ADA_lada} &\tiny\XSolidBrush& 5 & \underline{96.3} & \underline{97.7} & \underline{83.1} & 99.6 & \underline{85.0} & 99.7 & \underline{93.6}\\
    \cdashlinelr{1-10}
    AADA~\cite{su2020ADA} &\tiny\XSolidBrush& 10 & 93.5 &93.1 &83.2 &99.7 &84.2 &\bf{\underline{100.0}} &92.3 \\
    TQS~\cite{fu2021ADA} &\tiny\XSolidBrush& 10 & 96.4 &96.4 & 86.4 &\bf{\underline{100.0}} & 87.1 &\bf{\underline{100.0}} & 94.4\\
    CLUE~\cite{prabhu2021active} &\tiny\XSolidBrush& 10 & 96.2 & 94.7 & 84.4 & 99.4 & 81.0 & \bf{\underline{100.0}} & 92.6 \\
    SDM-AG~\cite{Xie2022CVPR} &\tiny\XSolidBrush& 10 & 95.9 & 96.4 & 86.1 & \bf{\underline{100.0}} & 86.5 & \bf{\underline{100.0}} & 94.2\\
    LADA~\cite{sun2022ADA_lada} &\tiny\XSolidBrush& 10 & \underline{97.8} & \underline{99.1} & \underline{87.3} & 99.9 & \underline{\textbf{87.6}} & 99.7 & \underline{95.2}\\
\midrule
    MHPL~\cite{wang2023mhpl} & \checkmark& 5 & 97.8 & 96.7 & 82.5 & 99.3 & \underline{82.6} & \textbf{\underline{100.0}} & 93.2 \\
    LFTL & \checkmark& 5 & \underline{98.0} & \underline{98.5} & \underline{82.6} & \underline{99.9} & 82.2 &\textbf{\underline{100.0}} & {\underline{93.5}} \\
    \cdashlinelr{1-10}
    LFTL & \checkmark& 10 & {\bf{98.9}} & \bf{99.4} & {\bf{87.8}} & \textbf{{100.0}} & {86.3} & \textbf{{100.0}} & \textbf{{95.4}}\\
\bottomrule

  \end{tabular}
\end{adjustbox}
 \end{minipage}
  \hfill %
   \begin{minipage}[]{0.43\linewidth}
    \includegraphics[width=1\textwidth]{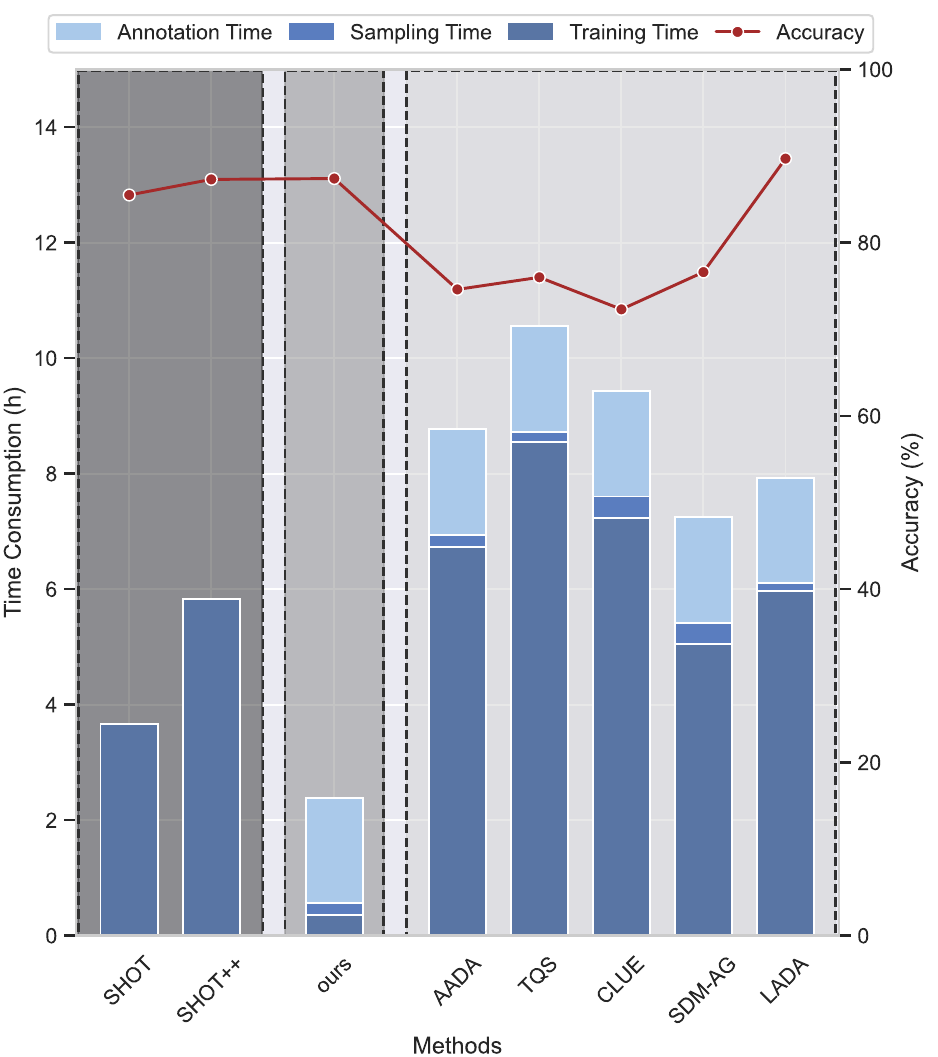}
    \vspace{-7mm}
    \caption{Comparison on the overall time consumption (training, active sampling and annotation time) and adaptation accuracy with SFUDA (SHOT) and ADA (AADA, TQS, CLUE, SDM-AG, LADA) methods on the VisDA-C dataset.}
    \label{fig:time_acc}
  \end{minipage}
\end{figure*}

\subsection{Comparison with SOTAs}
\label{sec:exp-comp}
We consider the following competitors: 
SFUDA methods including SFDA~\cite{kim2020sfda}, A$^2$Net~\cite{xia2021sfuda}, SHOT~\cite{liang2021sfuda}, SHOT$++$~\cite{liang2021sfuda}, CPGA~\cite{qiu2021CPGA}, DaC~\cite{zhang2022dac} and SF(DA)$^2$~\cite{hwang2024sfda2},
ADA methods including AADA~\cite{su2020ADA}, TQS~\cite{fu2021ADA}, CLUE~\cite{prabhu2021active}, SDM-AG~\cite{Xie2022CVPR} and LADA~\cite{sun2022ADA_lada},
and SFADA prior work MHPL~\cite{wang2023mhpl}
For \textit{fair inter- and intra-setting comparison}, consistent experimental conditions provided by SHOT are utilized for ADA methods based on the official codes of TQS~\cite{fu2021ADA}, CLUE~\cite{prabhu2021active}  and SDM-AG~\cite{Xie2022CVPR}, CLUE~\cite{sun2022ADA_lada} as well as the re-implementation of AADA~\cite{su2020ADA} provided by CLUE~\cite{prabhu2021active}. Results not reported by MHPL and not reproducible due to unavailability of source code are indicated by a dash.

We summarize the adaption accuracies on VisDA-C, Office-Home and Office-31 in Tab.~\ref{tab:visdac}, \ref{tab:officehome} and \ref{tab:office31}, respectively. 
On the large-scale dataset VisDA-C, given merely 1\% label annotations, LFTL is already capable of outperforming most of ADA methods. When the budget increased to 5\%, it exceeded the SFADA SOTA by 1.5\% and also outdid the source-accessible ADA SOTA under the same conditions by 0.7\%.
On the medium-sized Office-Home dataset, LFTL elevates the adaptation performance to a level of 78.8\% with 5\% annotation costs, which is comparable with SFUDA methods. We note a continuous enhancement in performance as the budget is raised to 10\%, culminating in the highest 85.1\% accuracy. 
Similar trend can also be observed in the small-sized Office-31 dataset. When 5\% target samples are labeled, our results surpass MHPL in average performance and in four out of five subtasks, while being on par with the source-accessible LADA.

The results show that, given a minimum amount of annotations, noticeable improvements can be achieved over SFUDA with increased flexibility. 
Meantime, despite the inavailability of the source domain, learning from what the model already learnt can offset the absence of the source data via our proposed CAS strategy and VPA approach.

\subsection{Efficiency Analysis}
\begin{figure*}[!t] %
  \begin{minipage}[]{0.46\linewidth}
\captionof{table}{Comparison on complexity and actual query time with ADA methods. Notations are explained in text.}
\label{tab:complexity}
\begin{adjustbox}{max width=1\textwidth}
  \newcolumntype{g}{>{\columncolor{mygray}}c}
   \begin{tabular}{lcc}
    \toprule
    Query Method & Complexity & Time (s/cycle) \\
    \midrule
    AADA~\cite{su2020ADA}   & $\mathcal{O}(NC + N\log N)$ & 79.16\\
    TQS~\cite{fu2021ADA}    & $\mathcal{O}(NCM + N\log N)$ & 127.51 \\ 
    CLUE~\cite{prabhu2021active} & $\mathcal{O}(tNBD)$ & 196.92\\
    SDM-AG~\cite{Xie2022CVPR} & $\mathcal{O}(NCD + N\log N)$ & 127.44 \\
    LADA~\cite{sun2022ADA_lada}& $\mathcal{O}(ND + N\log N)$ & 97.6 \\
    \midrule
    LFTL & $\mathcal{O}(NC + N\log N)$ & 73.96\\
    \bottomrule
    \end{tabular}
\end{adjustbox}
  \captionof{table}{Results of LADA minus source loss (SF-LADA) and our LFTL plus a source cross-entropy loss (S-LFTL). }
\label{tab:sflada}
\begin{adjustbox}{max width=1\textwidth}
    \begin{tabular}{c|c|c|c|c}
    \toprule
        Method & SF & 1\% VisDA-C & 5\% Office-Home & 5\% Office-31 \\
    \midrule
        LADA & \tiny\XSolidBrush & \underline{89.7} & 80.0 & 93.6\\
        S-LFTL & \tiny\XSolidBrush & 87.6 & \underline{80.1} & \underline{94.1} \\
    \midrule
        SF-LADA & \checkmark & 82.6 & 72.4 & 89.0 \\
        LFTL &\checkmark & \underline{87.4} & \underline{78.8}& \underline{93.5} \\
    \bottomrule
    \end{tabular}
\end{adjustbox}
\end{minipage}
  \hfill %
   \begin{minipage}[]{0.52\linewidth}
\includegraphics[width=0.98\linewidth]{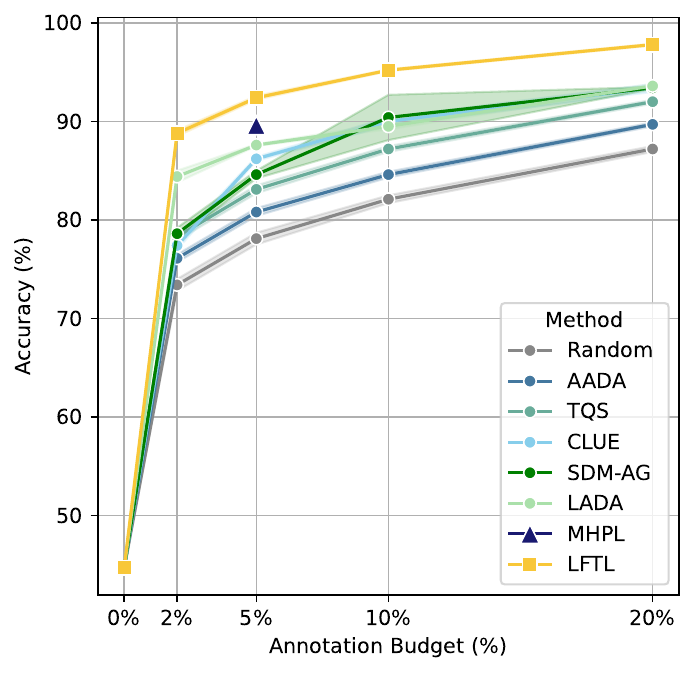}
\caption{Results of SFADA (MHPL and LFTL) and ADA (others) methods on VisDA-C (ResNet50) as the annotation budget grows.}
\label{fig:largestride}
  \end{minipage}
\end{figure*}

We compare with open-source SFUDA and ADA methods in terms of actual time consumption, including model training, active sampling and human annotation, as presented in Fig.~\ref{fig:time_acc}. We take the adaptation on the VisDA-C dataset for example, and provide 1\% budget, \ie 550 images of the target domain for active learning-based methods. Note that here we report the consumption of the whole sample-and-adaptation process, instead of the per-round time. All experiments were conducted on 1 NVIDIA GeForce RTX 3090 GPU, with 80 CPU cores and 256 GB memory. 

Considering the \textbf{training time for adaptation}, SFUDA SOTAs often require sophisticated schemes (clustering, two-stage training, \etc) to adapt to a completely unlabeled domain, which necessitate more computational resource. For example, SHOT and SHOT++ takes 3.7h and 5.8h to converge respectively, and CPGA uses more than five days to attain the reported performance, which we omit in Fig.~\ref{fig:time_acc} to prevent it from dominating the figure.
On the other hand, when equipped with both source and target data, ADA methods are expected to directly align different domains, as well as identifying novel samples, which adds to the training burden. For example, AADA, CLUE and TQS consume more than 6h mainly because of the adversarial training. TQS additionally trains five classifiers to construct a voting committee for active learning, culminating in an overall duration of 8.5h. LADA employs local neighbor search for more reliable training guidance, which takes 5.9h. SDM-AG improves on the optimization complexity yet still requires 5.1h. 
In contrast, drawing upon queried informative data, our LFTL leverages the iterative nature of active learning instead of additional computations. The proposed VPA retrospects on the knowledge gained from the source and intermediate models of previous rounds, which is temporarily ensembled in the VP vault for readily use. As the comparison shows, it significantly outperforms competitors with only 0.3h for adaptation while yieding superior accuracy.

As for the \textbf{active sampling efficiency}, we summarize the theoretical complexity and actual training duration for active methods in Tab.~\ref{tab:complexity}, where $C$ and $N$ to denote the number of classes and unlabeled instances respectively. 
AADA and our LFTL operate with $\mathcal{O}(NC + N\log N)$ complexity, which are the most efficient approaches. 
The clustering-based CLUE works at $\mathcal{O}(tNBD)$, in which $D$ denotes the embedding dimension (512), $B$ is the budget, and $t$ is the number of clustering iterations (300). 
The ranking-based TQS and SDM-AG uses $\mathcal{O}(NCM + N\log N)$ and $\mathcal{O}(NCD + N\log N)$ respectively, in which $M$ represents the number of members in the classifier committee (5 is adopted). The local neighbor search in LADA leads to a complexity of $\mathcal{O}(ND + N\log N)$. The feature dimension $D$ of SDM-AG and LADA is 256.
Contrary to previous active criteria designed for sampling in the target domain, CAS contrasts the hypotheses generated by models from successive rounds to identify target samples that deviate from the source domain and persistently challenging for the recognition task.
The comparison shows that the proposed CAS is effective, scalable and efficient.

Regarding the \textbf{human annotation time}, given that image-level labels take 1 second per class~\cite{papadopoulos2014training,bearman2016u},  1\% target domain subset of VisDA-C would cost $550\times 12 \approx 1.83$h. 
As Fig.~\ref{fig:time_acc} shows, compared to non-active methods, although factoring in the annotation time introduces additional burden for ADA methods, we still retain the top place.

Collectively, the comparative analysis of accuracy and efficiency reveals that our LFTL framework, through the synergy of CAS and VPA strategies, can be freed from source-accessibility, specialized architectures or complex optimization techniques while still delivering SOTA performance.

\subsection{Promises of Continual Performance Growing}
\label{sec:largerbuget}
In addition to the experiments presented in main results under varying budget constraints, in this section, we explore the trade-off between the cost and adaptation performance with increased budget and larger strides. The budget is increased to 2\%, 5\%, 10\% and 20\% successively, with the corresponding results summarized in Figure~\ref{fig:largestride}.
We observe that the proposed LFTL framework performs notably better than competitors in terms of both accuracy and robustness. 
The performance superiority is consistent across different budget scenarios, from tight to generous, indicating its effectiveness in leveraging data resources. 
On the contrary, other methods either gradually deviate from the source domain throughout the process, failing to acquire domain-variant novel knowledge, or they solely support one-round active sampling as in MHPL.
We emphasize that the promise of continual performance growth holds practical value in real-world applications, as it facilitates the effective feedback loop between data reflow and model iteration.

\subsection{Discussion of Source Availability}
Considering the performance margin of SFADA over ADA, we take LADA, the ADA SOTA, as an example for comparison to explore the role of source data in adaptation.
We first ablate the source domain loss from LADA, denoted as SF-LADA to compare with LFTL.
Given that the source data becomes unavailable, each epoch now iterates over the labeled target dataset instead of the labeled source dataset, the same as our LFTL.
The averaged results in Tab.~\ref{tab:sflada} show that, without specifically designed hypothesis transfer strategy, removing the source data from ADA methods will cause task models to forget the previously learnt knowledge in the source domain during adaptation. 
And the newly acquired information from the target domain is too limited to counteract it, resulting in noticeable accuracy gaps. 
This further validates the necessity  of learning from the learnt under the source-free condition.

Then we add a simple cross-entropy loss for the source data to LFTL, namely S-LFTL, while keeping the number of iterations unchanged. As shown in Tab.~\ref{tab:sflada}, S-LFTL improves over LFTL, demonstrating that direct source supervision could still provide more useful information than hypothesis transfer and memory recollection. 
The source data proves especially effective for smaller Office datasets, where the improvements are noticeable while the additional amount of additional time is affordable.
For the large dataset, although the performance of S-LFTL does not match that of LADA within just 6\% of its adaptation time, extending the training iterations to access all available data would guarantee further performance improvements.

\begin{table}[t!]
\centering
\caption{Comparison between the proposed CAS and active learning baselines in terms of classification accuracy ($\%$). Experiments are performed on the large-scale VisDA-C dataset (ResNet101) with 1\% budget in total for 10 cycles. The best results are highlighted in bold.}
\label{tab:visdac-actives}
\setlength{\tabcolsep}{1.0mm}{
\begin{adjustbox}{max width=1\textwidth}
  \scalebox{0.8}{
  \newcolumntype{g}{>{\columncolor{mygray}}c}
  \begin{tabular}{lcccccccccc}
    \toprule
    Method & 1 & 2 & 3 & 4 & 5 & 6 & 7 & 8 & 9 & 10 \\
    \midrule

    Random & 67.0 & 70.6 & 73.7 & 76.6 & 78.7 & 79.9 & 81.1 & 81.8 & 82.2 & 82.7 \\
    Ent-max & 62.2 & 67.9 & 73.7 & 77.6 & 79.9 & 80.5 & 80.2 & 80.2 & 80.0 & 80.5 \\
    Ent-min & 60.4 & 66.4 & 64.5 & 64.2 & 63.8 & 64.0 & 64.1 & 63.5 & 64.6 & 63.9 \\
    Kcenter-greedy & {67.5} & 71.3 & 74.5 & 77.4 & 78.9 & 80.6 & 81.4 & 82.3 & 82.7 & 83.5 \\
    Kmeans & 60.5 & 67.3 & 72.9 & 75.7 & 77.5 & 78.7 & 79.7 & 80.4 & 80.9 & 81.2 \\
    Least-confidence & 63.2 & 69.9 & 74.5 & 78.5 & 80.2 & 80.9 & 80.7 & 81.1 & 81.2 & 81.9 \\
    Bayesian & 66.5 & 70.4 & 73.9 & 76.7 & 78.7 & 80.2 & 81.4 & 82.1 & 82.8 & 83.4 \\
    BvSB & 65.7 & {72.2} & {77.0} & 79.8 & 81.4 & 82.5 & 83.4 & 84.0 & 84.6 & 85.1 \\
    \midrule
    
    LFTL & \textbf{67.6} & \textbf{74.9} & \textbf{79.1} & \textbf{81.5} & \textbf{83.6} & \textbf{84.3} & \textbf{85.6} & \textbf{86.2} & \textbf{86.9} & \textbf{87.4} \\

    \bottomrule
  \end{tabular}
  }
\end{adjustbox}
}
\end{table}

\subsection{Comparison with Active Learning Baselines} 
To validate the proposed active  query strategy, we compare LFTL with active learning baselines. We choose Random, uncertainty-based EntMax~\cite{settles2009activesurveyentropy}, Least Confidence~\cite{lewis1994leastconf}, BvSB~\cite{lewis1994leastconf}, Bayesian~\cite{gal2017bayesian}, distribution-based K-means~\cite{yan2009kmeans} and K-center greedy~\cite{sener2018coreset}. 
Experiments are performed on the large-scale VisDA-C dataset with 1\% annotation budget in total for 10 rounds.
The results for each round are listed in Tab.~\ref{tab:visdac-actives}.
It manifests that the proposed sampling strategy consistently outperforms its substitutes.
As the iteration proceeds, information queried by other active learning methods become redundant, failing to fulfill the requirements of the DA task. In comparison, 
the CAS and class-level transferability estimation we propose efficiently prioritize samples with local uncertainty, global intransferability and temporal stagnancy along the sample-and-adaptation progress, bringing continuous growth to the DA task.

\section{Conclusion}
In this paper, we investigate the challenging source-free active domain adaptation (SFADA) setting, where source data becomes inaccessible during adaptation, and a minimum amount of annotation budget is available for the target domain. 
We present a shared framework for both active sampling and domain adaptation to learn from the hypotheses and feature representations of the learnt source model and actively iterated models. 
The proposed CAS criterion effectively prioritizes samples that are both informative to the current model and persistently challenging throughout the iterative process, and their visual understandings of actively queried samples are temporally preserved and exploited via the VPA strategy during adaptation.

Extensive experiments on three DA benchmarks have shown that, in comparison with ADA and SFUDA, SFADA provides a worthy trade-off between annotation costs and model performance in both accuracy and efficiency. In comparison with previous SFADA competitors, our LFTL framework also presents superior performance and exhibits more flexibility than the one-round method.

\section*{Acknowledgment}
This work was supported by National Science and Technology Major (2021ZD0114703-2), National Natural Science Foundation of China (Nos. 61925107, 62271281, 62021002).

\bibliographystyle{splncs04}
\bibliography{main}

\appendix
\clearpage
\setcounter{page}{1}
\maketitlesupplementary

\setcounter{section}{0}
\renewcommand\thesection{\Alph{section}}

\begin{figure}
     \centering
     \includegraphics[width=1\linewidth]{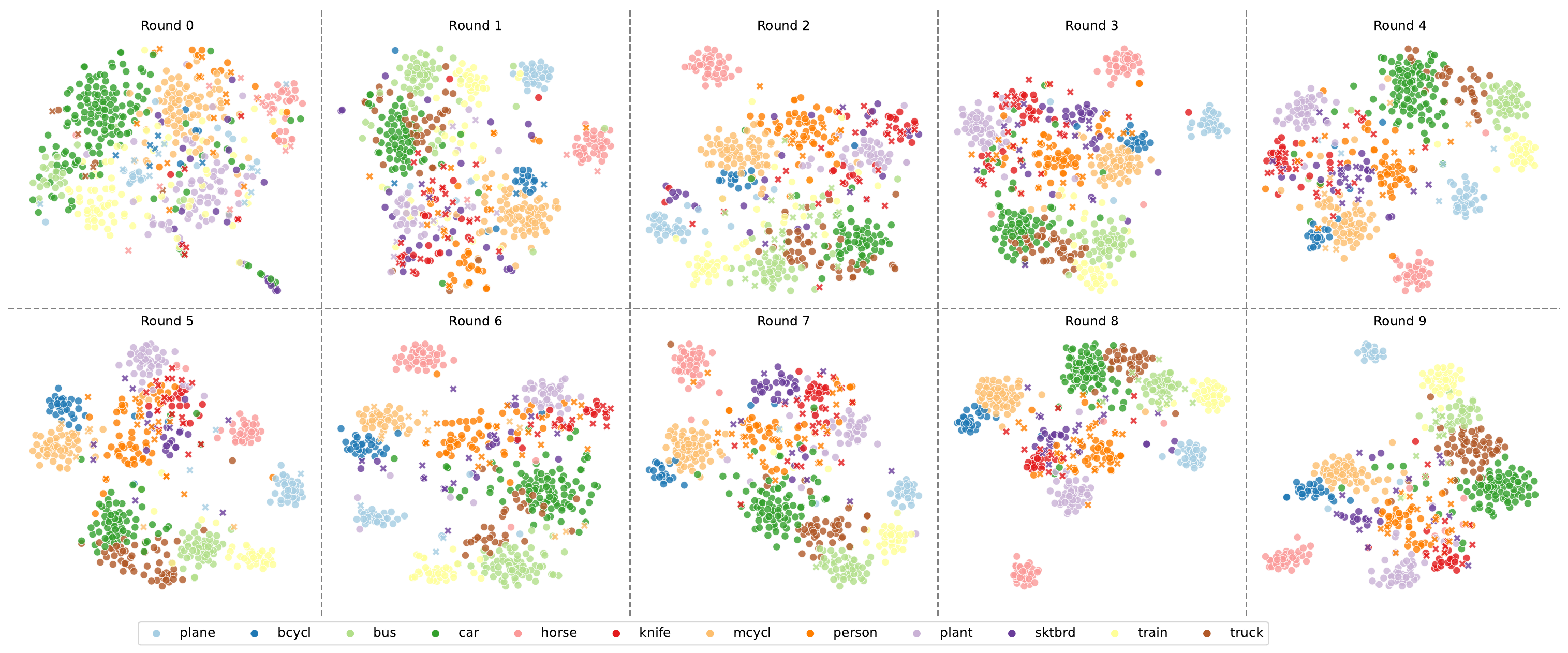}
      \captionof{figure}{t-SNE visualization of unlabeled target samples (colored by classes) and actively queried samples (marked by cross) on VisDA-C with 0.1\% budget per round.}
     \label{fig:active_rounds}

\end{figure}

\setcounter{page}{1}

\section{Qualitative Analysis}

\subsection{t-SNE Visualization}
To visualize the task model's comprehension of the target data and the selection of informative samples during the iterative process of active learning and domain adaptation, we present the t-SNE~\cite{van2008tsne} plots for the VisDA-C dataset in Fig.~\ref{fig:active_rounds}. For the sake of visual clarity, only a subset of unlabelled data is randomly sampled and depicted in each round.

We first observe that, as the volume of actively selected samples increases, the classification boundaries are progressively revealed, indicating that the task model's understanding of the target data distribution has become more comprehensive. 
As our CAS is motivated, under the domain shift problem, unlabeled samples we identify as challenging to transfer are proximate to decision boundaries in the target domain, where our active selection often happens. At the same time, our sampling process prioritizes minority classes (\eg knife), which is contributed to our class-level transferability estimation. 

\subsection{Queried Target Samples}
\begin{figure}
    \centering
    \includegraphics[width=\linewidth]{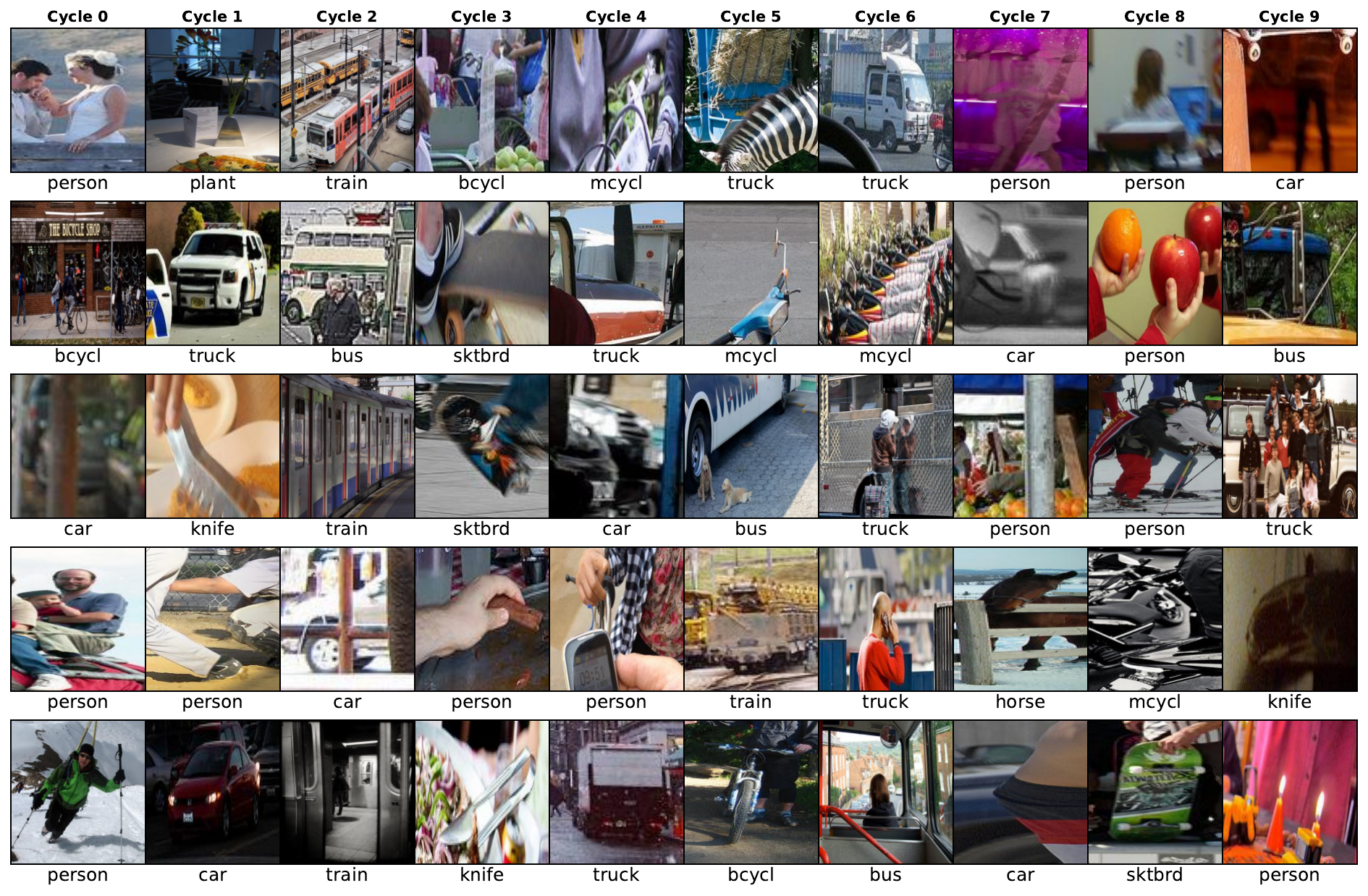}
    \caption{Top-5 CAS samples of 10 rounds on VisDA-C.}
    \label{fig:activeselection}
\end{figure}
We further validate the efficacy of proposed CAS via visual inspection of the queried target samples. Fig.~\ref{fig:activeselection} shows top-5 actively queried samples of each active learning round on VisDA-C. 
Based on the recognition results from the source-only model, person (19.2\%), truck (6.1\%) and knife (4.9\%) are challenging classes. We can observe that more iconic and representative samples are selected in order to learn authentic features that differ from those of the generated images. On the other hand, for more transferable classes, \eg car (80.0\%) and motorcycle, samples that are truncated, occluded, distant and blurry are queried at first. Those patterns are not present in the source domain but are ubiquitous in real world scenarios, which provide rich information for the adaptation. 

We have also noted that in the later stages of iteration, our CAS not only selects samples that are prone to confusion but is also able to identify instances with inaccurate labels. For example, the image where human hands holding apples is labeled as person, or the case where the foreground is a person but the label corresponds to a blurry truck in the background.

\section{Component Verification}
\subsection{Ablation Study}
Take the combination of vannila BvSB~\cite{lewis1994leastconf} sampling strategy and labeled target supervision via CE loss as a baseline, we present the ablation study of the propose LFTL framework in Tab.~\ref{tab:abl}.
On VisDA-C and Office-31 datasets with differing budget constraints, consistent improvements can be observed with each component, which validate our motivations. 
Given the same annotation budget, our CAS strategy prioritizes target samples that remain unrecognizable to the current model and have not been captured in preceding active learning rounds, and meantime it factors out samples with knowledge previously acquired when the hypotheses exhibit increased confidence. In addition to the contrastive margin $u_{cm}$, the class-level transferability $u_{ct}$ enhances our sampling criterion with a global semantic perspective, which promotes the selection of tail and challenging classes. 
The actively selected target samples then play the role of anchors to guide the optimization during the adaptation procedure. Without explicit error-prone pseudo-labeling or time-consuming clustering, the $\mathcal{L}_{ac}$ efficiently encourages density around active anchors. When replaced with features with VP delivered via $\mathcal{L}_{VPA}$, the representations of active anchors are features temporally ensembled from the source model to the current, which efficiently promotes a source-like feature distribution in the target domain.

\begin{table}[!t]
\caption{Ablation study with 1\%-labeled VisDA-C and 5\%, 10\%-labeled Office-31. 
}
\centering
\begin{tabular}{cccc|c|cc}
    \toprule
         $u_{cm}$ & $u_{ct}$ &  $\mathcal{L}_{ac}$ & $\mathcal{L}_{\tiny{vpa}}$ & 1\%-VisDA-C & 5\%-O31 &  10\%-O31 \\ 	
    \midrule
    & & &  & 
    85.1 & 89.2  & 93.1 \\
    $\checkmark$ & & & & 
    86.7& 91.4 & 94.8 \\ 
    $\checkmark$ & $\checkmark$ & &   & 
    86.9 & 91.8 & 95.1 \\ 
    $\checkmark$ & $\checkmark$ & $\checkmark$ &  & 
    87.0 &92.0 & 95.3 \\
    $\checkmark$ & $\checkmark$ & & $\checkmark$  & 
    \bf{87.4} &\bf{93.5} & \bf{95.4} \\ 
    \bottomrule
\end{tabular} 
\label{tab:abl}
\end{table}

\subsection{CAS for SFUDA SOTAs}
This section further verifies CAS by applying it to SFUDA methods to facilitate their adaptation.
Here we take SOTAs SHOT and SF(DA)$^2$ for example. Results in Tab.~\ref{tab:sfudacas} show that CAS can be simply plugged into source-free methods to query informative target samples for a performance boost. 
Meantime, a better approach is to simultaneously consider the risk of forgetting the domain-invariant knowledge  during transfer and thereby incorporating our VPA strategy to learn from the learnt.

\begin{table}[t]%
\caption{Results of applying CAS to SFUDA SOTAs on VisDA-C.}
\centering
\setlength{\tabcolsep}{3pt}
    \begin{tabular}{c|c|c|c|c|c|c}
    \toprule
        AS & CAS & SHOT & SHOT++ & SF(DA)$^2$ & MHPL &  LFTL \\
    \midrule
        0 & \tiny\XSolidBrush & 85.5 & 87.3 & \underline{88.1} & - & -\\
        1 & \checkmark & 86.4 & 87.5 & \underline{88.9} & -& 87.4\\
        5 & \checkmark & 91.9 & 91.8 & 92.3 & 91.3 & \underline{92.8}\\
        
    \bottomrule
    \end{tabular}
\label{tab:sfudacas}
\end{table}

\subsection{Parameter Sensitivity Analysis}
\begin{figure}[t]
    \centering
    \includegraphics[width=\linewidth]{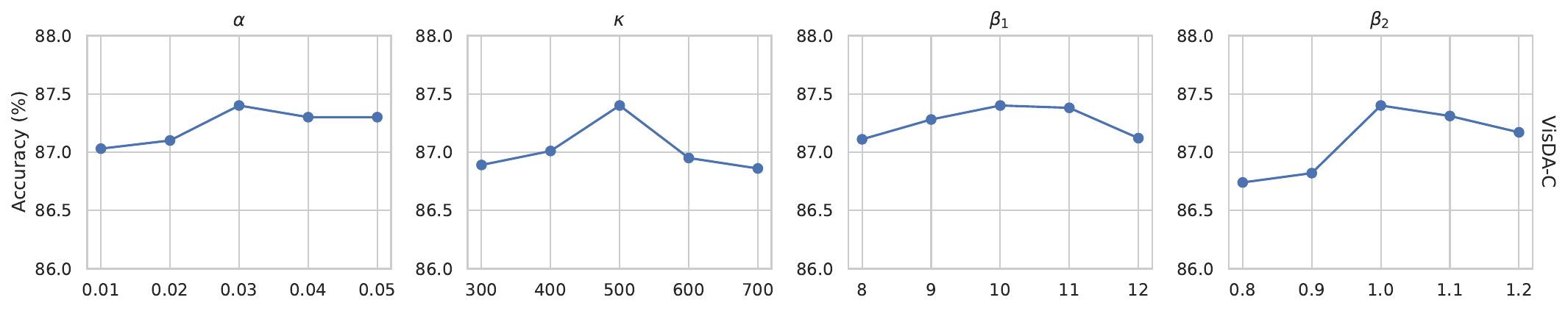}
    \caption{Parameter sensitivity analysis of $\alpha$ in Eq.~1,  $\kappa$ in Eq.~3, and $\beta_1$, $\beta_2$ in Eq.~10 on VisDA-C.}
    \label{fig:param_sensitivity}
\end{figure}
\label{supsec:sensitivity}

To validate the effectiveness and generalization ability of the proposed method, we study the sensitivity of LFTL to $\alpha$ in Eq.~1,  $\kappa$ in Eq.~3, and $\beta_1$, $\beta_2$ in Eq.~10 on the VisDA-C dataset.
We experiment around the optimal values of parameters, perform three trials with a set of seeds and average the results. 
In Fig.~\ref{fig:param_sensitivity},
we observe similar bell-shaped curves on all experiments, indicating consistent performance gains benefit from our proposed methods and demonstrating robustness to parameter choices.

\section{Additional Implementation Details}
During the sample-and-adaptation interplay, the SGD optimizer with a momentum of 0.9 and a weight decay of $1e{-3}$ is applied. 
The learning rate is set as $1e{-3}$ for VisDA-C and $1e{-2}$ for Office, scheduled by $\eta =\eta_0(1+10p)^{-0.75}$ where $p$ increases from 0 to 1 during optimization. We set batch size as 64. 
For all datasets, the CAS coefficient $\alpha=0.03$, $\beta_1, \beta_2$ 
 in model adaptation are set to 10.0 and 0.9 to balance the magnitude of each loss terms. We set $\kappa=500$ for the large-scale VisDA-C and 100 for smaller Office datasets.


\end{document}